\documentclass[journal]{IEEEtran}
\usepackage{ifpdf}
\usepackage{float}
\usepackage{cite}
\usepackage[cmex10]{amsmath}
\usepackage{algorithmic}
\usepackage{array}
\usepackage{mdwmath}
\usepackage{mdwtab}
\ifCLASSINFOpdf
   \usepackage[pdftex]{graphicx}
   \graphicspath{{../pdf/}{../jpeg/}}

   \DeclareGraphicsExtensions{.pdf,.jpeg,.png}
\else

   \usepackage[dvips]{graphicx}

  \graphicspath{{../eps/}}

   \DeclareGraphicsExtensions{.eps}
\fi

\ifCLASSINFOpdf

\else

\fi

\begin{document}
%
\title{Facial Surface Analysis using Iso-Geodesic Curves in Three Dimensional Face Recognition System }
%
%
%

\author{Rachid AHDID,
        El Mahdi BARRAH,
        Said SAFI
        and~Bouzid MANAUT
\thanks{Rachid AHDID, El Mahdi BARRAH and Said SAFI are with the Department of Mathematics and Informatics, Sultan Moulay Slimane University, Beni Mellal, Morocco, e-mail: (r.ahdid@usms.ma).}
\thanks{Bouzid MANAUT was with Department of physics  Sultan Moulay Slimane University, Beni Mellal, Morocco.}
\thanks{Manuscript received xxxx; revised xxxx.}}

\maketitle

\begin{abstract}
In this paper, we present an automatic 3D face recognition system. This system is based on the representation of human faces surfaces as collections of Iso-Geodesic Curves (IGC) using 3D Fast Marching algorithm. To compare two facial surfaces, we compute a geodesic distance between a pair of facial curves using a Riemannian geometry. In the classifying step, we use: Neural Networks (NN), K-Nearest Neighbor (KNN) and Support Vector Machines (SVM). To test this method and evaluate its performance, a simulation series of experiments were performed on 3D Shape REtrieval Contest 2008 database (SHREC2008).\\


\end{abstract}

\IEEEpeerreviewmaketitle

\section{Introduction}
Face recognition is one of the most commonly used techniques in biometrics authentication applications, of access and video surveillance control, this is due to its advantageous features. In a face recognition system, the individual is subject to a varied contrast and brightness lighting, background. This three-dimensional shape, when it is part of a two-dimensional surface, as is the case of an image, can lead to significant variations. The human face is an object of three-dimensional nature. This object may be subject to various rotations not only flat but also space and also subject to deformations due to facial expressions. The shape and characteristics of this object also change over time.\\

In these last years, a number of methods have been proposed for the recognition of human faces. In spite of the results obtained in this domain, the automatic face recognition stays one of very difficult problem. Several methods were developed for 2D face recognition. However, it has a certain number of limitations related to the orientation of the face or laying, lighting, facial expression. These last years, we talk more and more about 3D face recognition technology as solution of 2D face recognition problems. There are methods of 3D face recognition based on the use of three-dimensional information of the human face in the 3D space. Existing approaches that address the problem of 3D face recognition can be classified into several categories of approaches: Geometric or Local approaches 3D, Bronstein et al [1, 2] propose a new representation based on the isometric nature of the facial surface, Samir et al [3, 4] use 2D and 3D facial curves for analyzing the facial surface; Holistic approaches, Heseltine et al [5] have developed two approaches to applying the representations PCA Three-dimensional face, Cook et al [6] present a robust method for facial expression based on Log-Gabor models from images of deep and some approaches are based on face  segmentation can be found in [7 , 8, 9, 10, 11, 12].\\

In this work we present an automatic 3D face recognition system based on facial surface analysis using a Riemannian geometry. For this we take the following steps:\\

- 	Detection of 3D face where the nose end is a reference point.\\

-	Iso-geodesic curves extraction using a 3D Fast Marching algorithm.\\

-	Compute of geodetic distance between two iso-geodesic curves using mathematical formulas in Riemannian metric.\\

The rest of this paper is organized as follows: Section 2 describes the methodology of the proposed method with its stages: reference point detection, geodesic distance computing, facial curves extraction. Section 3 includes the simulation results and analysis. Section 4 draws the conclusion of this work and possible points for future work.\\

\section{METHODOLOGY}
Given an image of 3D face Shape REtrieval Contest 2008 database (SHRED2008) database our goal is to realize an automatic 3D face recognition system based on the computation of the geodesic distance between the reference point (nose tip) and the other points in the 3D face surface. So, our algorithm is divided to four main steps, first: Reference Point Detection, in this paper we have detected the point of reference (nose tip) manually. Second: Geodesic Distance computation, an effective method to compute a geodesic distance between two points of facial surface is using the Fast Marching as a numerical algorithm for solving the Eikonal equation. Third: facial curves extraction and compute a geodesic distance between each pair of curves. Finally: classification algorithms, in this step we use three types of classification algorithms: the Neural Networks (NN), k-Nearest Neighbor (KNN) and Support Vector Machines (SVM). Figure (1) illustrates the steps of proposed method in our 3D face recognition system.\\
\begin{figure}[H]
\centering
\includegraphics[width=4in]{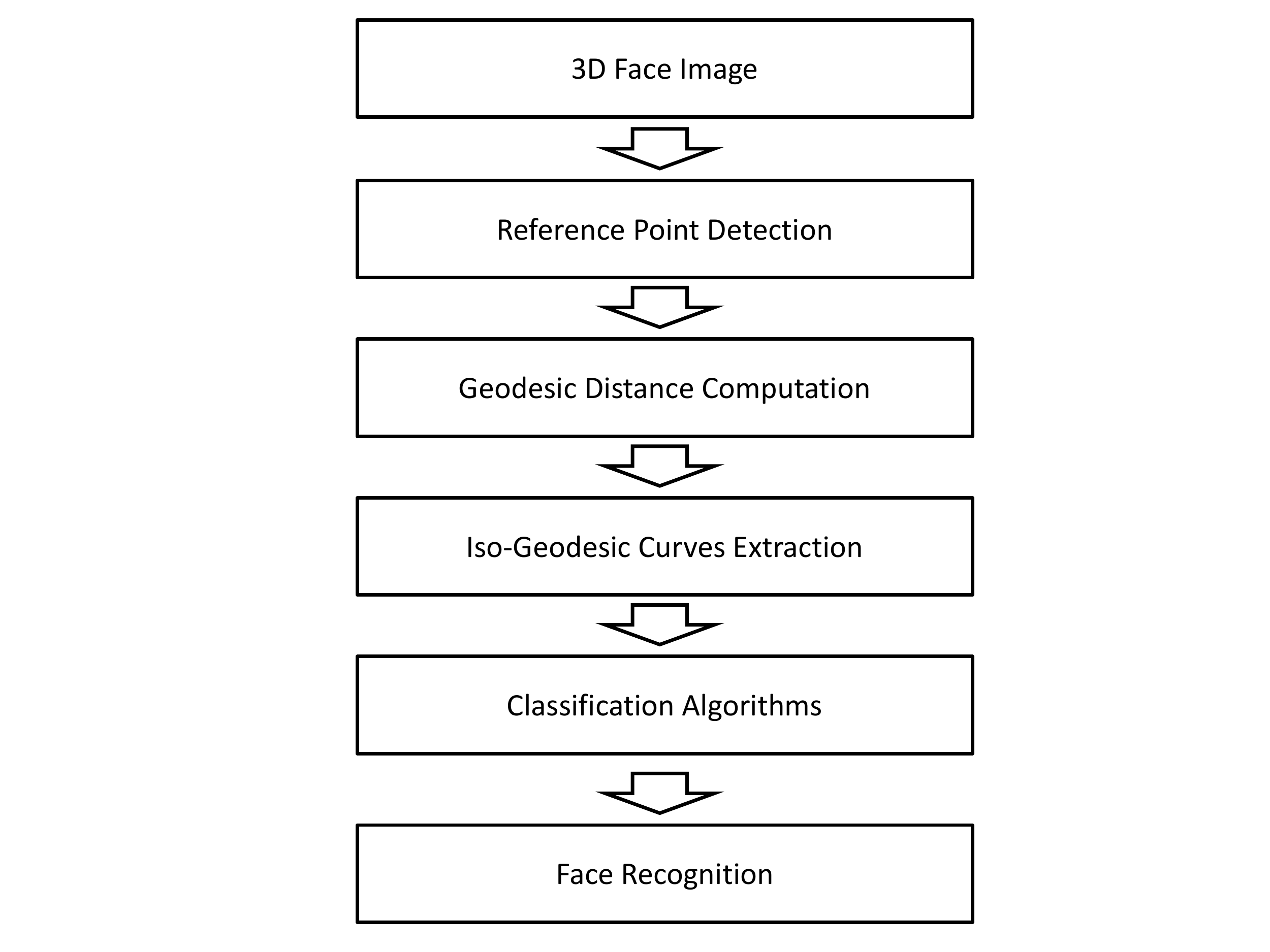}
\caption{Methodology Architecture}
\label{fig_sim}
\end{figure}

\subsection{Reference Point Detection}
The reference point (nose tip) is detected manually or automatically. There are several automatic approaches: L. Ballihi et al are developed an automatic algorithm nose end detection of a 3D face in [13]. This algorithm is based on two cuts of the facial surface. The first is at transverse face of the mass center. The second cut is based on the minimum depth point of the horizontal curve obtained by the first cut. The output of the last cut is a vertical curve and the minimum depth of this curve is the end of 3D face nose. In [14] S. Bahanbin et al use Gabor filters to automatically detect the nose tip. Another method has been used by C. Xu et al in 2004 [15], this method computes the effective energy of each neighbor pixel, then be determined the mean and variance of each neighbor pixel and uses the SVM to specify point end of the nose. L.H. Anuar et al [16] use a local geometry curvature and point signature to detect a nose tip region in 3D face model.\\

In this work we have detected the point of reference $p_{0}$ (nose tip) manually. The following figure (figure2) summarizes the steps to detect the nose tip of a 3D face an image of the database SHREC2008:
\begin{figure}[H]
\centering
\includegraphics[width=3in]{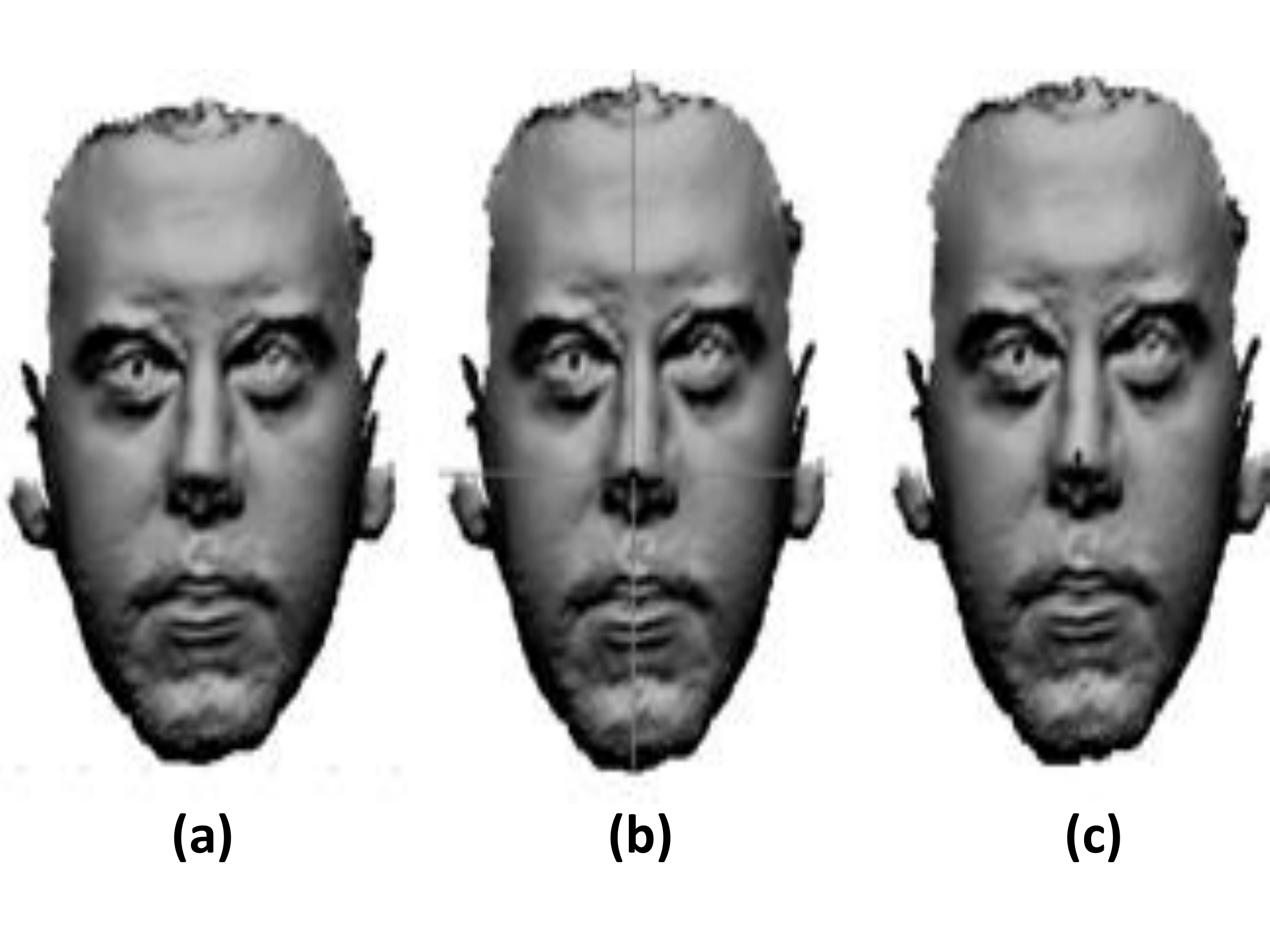}
\caption{3D face nose end detection steps: (a) 3D face image; (b) Manual nose tip selection; (c) Reference point detection}
\label{fig_sim}
\end{figure}

\subsection{Geodesic Distance}
The geodesic distance between two points’ p0 and p of 3D face surface is the shortest path between the two points while remaining on the facial surface. In the context of calculating the geodesic distance R. Kimmel and J.A. Sethian [17] propose the method of Fast Marching as a solution of the Eikonal equation.\\

The Eikonal equation given as:
\begin{equation}
   \mid \bigtriangledown_{u} (x) \mid = F(x); \quad x \in \Omega
\end{equation}
with:\\ - $\Omega$ is an open set in $R^{n}$ housebroken limit.\\ - $\bigtriangledown$ denotes the
gradient.\\ - $\mid . \mid$ is the Euclidean norm.\\

The Fast Marching method is a numerical method for solving boundary value problems of the Eikonal equation [17, 18, 19]. The algorithm is similar to the Dijkstra's algorithm [20] and uses that information flows only to the outside from the planting area.\\

We consider a 3D face surface discretized using a triangular mesh with N vertices. Figure (3) shows a 3D face image of the Shape REtrieval Contest 2008 (SHREC2008) database whose obverse surface is discretized into a triangular mesh.
\begin{figure}[H]
\centering
\includegraphics[width=3in]{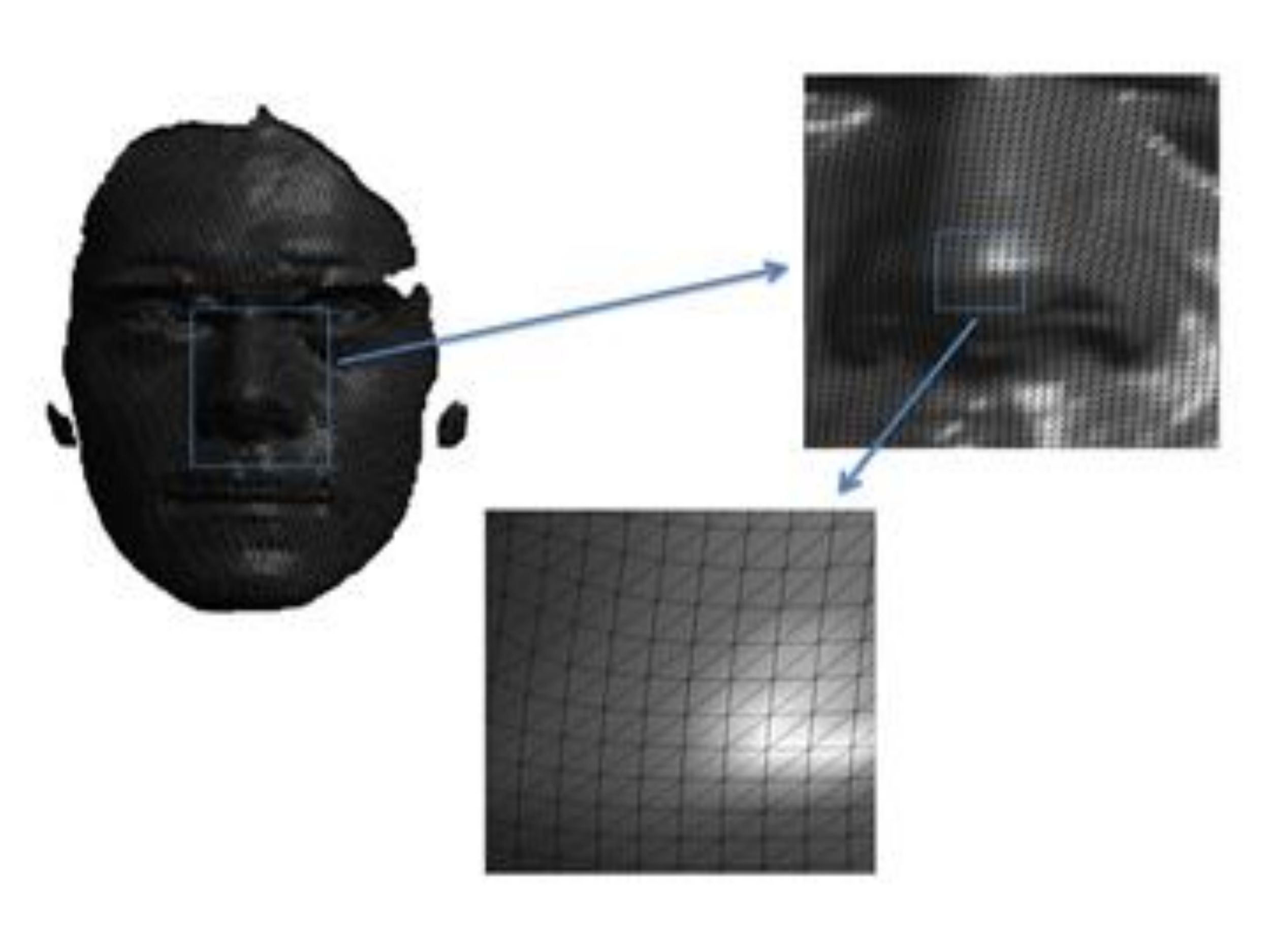}
\caption{3D facial surface discretized on triangular mesh of N vertices}
\label{fig_sim}
\end{figure}
The geodesic distance between two points on a surface is calculated as the length of the shortest path connecting the two points. Using the Fast Marching algorithm on the triangulated surface 3D face, we can compute the geodesic distance between the reference point P0 and the other point’s p constructing the facial surface.\\

The geodesic distance $\delta_{p_{0},p}$ between $p_0$ and p is approximated by the following expression:
\begin{equation}
   \delta_{p_{0},p} = min \gamma (\beta(p_{0},p))
\end{equation}
with:\\ - $\beta(p_{0},p)$ is the path between $p_0$ and according to the facial surface $S$ of the 2D face.\\ - $\gamma (\beta(p_{0},p))$ is the path length.\\

The following figure (Figure4) shows the steps for determining the geodesic distance using a 3D face image of SHREC2008 database.
\begin{figure}[H]
\centering
\includegraphics[width=3in]{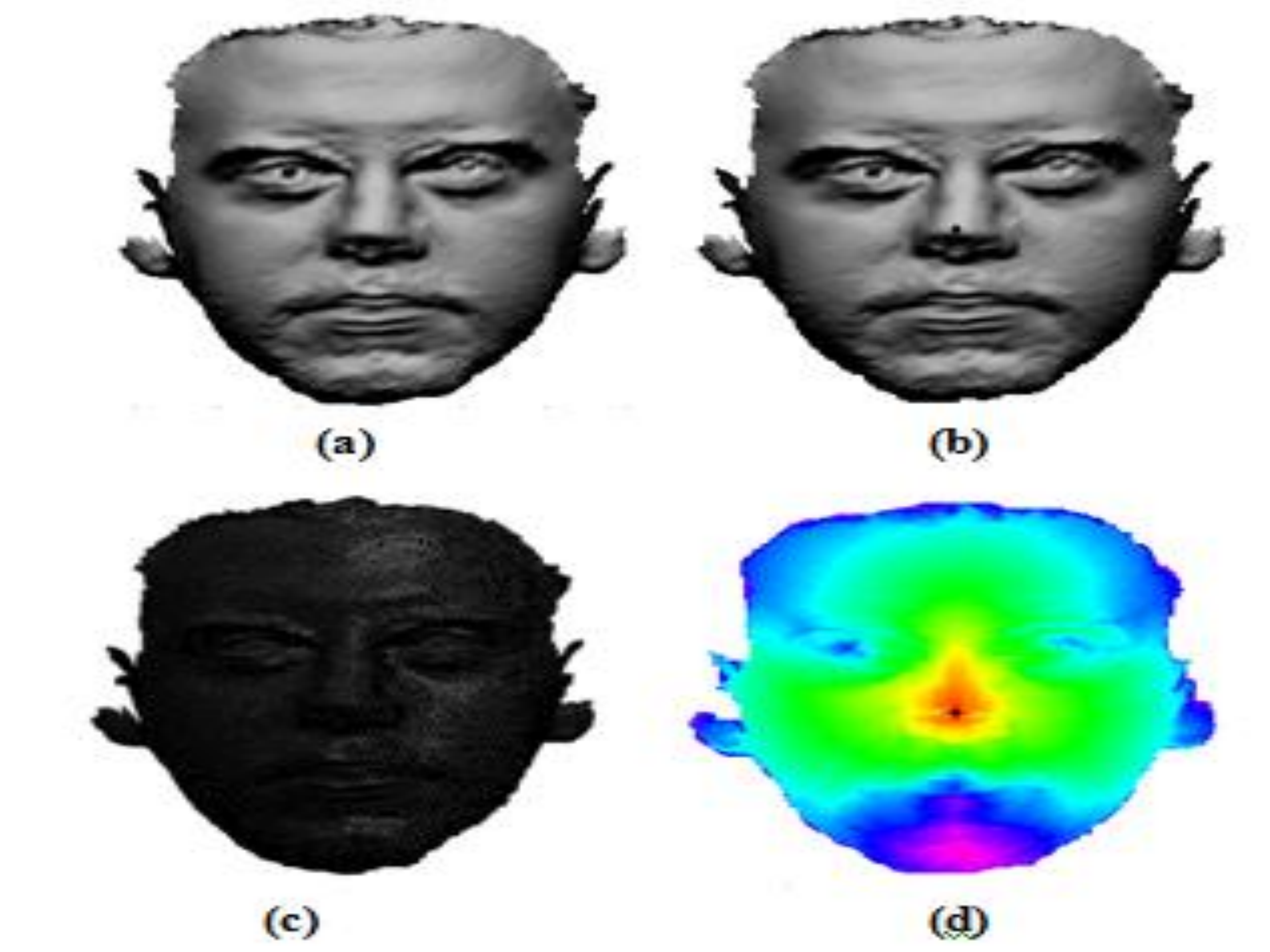}
\caption{3D face geodesic distance computes Steps: (a) 3D face image; (b) Reference point detection; (c) Discretization by triangular mesh; (d) Geodesic distance computing}
\label{fig_sim}
\end{figure}

\subsection{Facial Curves}
This 3D face recognition method is based on the analysis of facial surfaces by analyzing of facial curves using Riemannian geometry. To extract the curves of a 3D face surface, the first step is to define the real-valued function on this surface [17]. According to the extraction strategy, different types of facial curves can be found: 1- Iso-depth curves, these curves are obtained by the intersection of the 3D face surface with parallel planes perpendicular to the direction of watching. The depth curves located at equal values of z [17]. 2-Iso-radius curves, these curves are determined by the intersection of the facial surfaces with spheres as centre is the reference point of 3D face (nose tip) and variable radius [13]. 3- Iso-geodesic curves are defined as the locations of all points on the facial surface having the same geodesic distance to the reference point chosen (in our case the end of the nose). The geodesic distance between two points on a surface is the shortest path between these two points along of surface [13, 17].\\

In this work, we represent the 3D human face surface by a collection of iso-geodesic curves. To extract the iso-geodesic curves we use the Fast Marching algorithm as a solution of Eikonal equation [19].  Figure (5) present the steps of extracting of iso-geodesic curves in some 3D face images of SHREC2008 database.
\begin{figure}[H]
\centering
\includegraphics[width=3in]{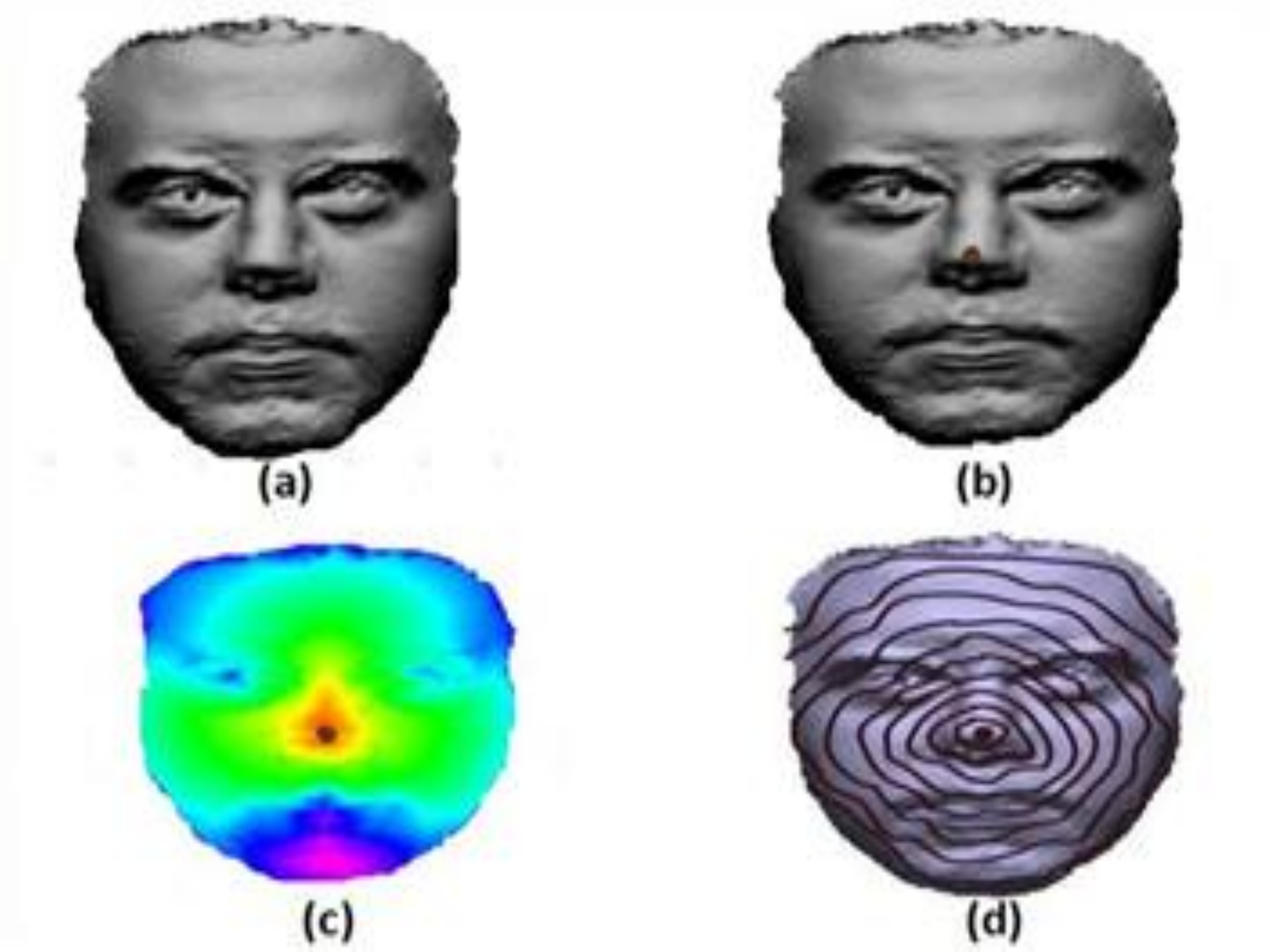}
\caption{Extracting facial curves steps: (a) Pre-treated face image, (b) Detection of the reference point, (c) Compute of geodesic distance, (d) Facial curves extraction}
\label{fig_sim}
\end{figure}
Given two points on a face surface S (reference point $p_0$ and $p$), A geodesic distance between $p_0$ and $p$ is defined as the arc length of the shortest path between these two points along the surface and denoted by a Geodesic Distance Function (GDF), which is a continuous function on the facial surface.

\begin{equation}
   F(p_{0},p) = k; \quad k \in [0, + \propto[ and (p_{0},p) \in S
\end{equation}

We can therefore define the facial curves by:
\begin{equation}
   c_k = \{ p \in S \setminus F(p_{0},p) = k \} \subset S; \quad k \in [0, + \propto[
\end{equation}

The function F defines the geodesic distance between $p_0$ and $p$, or the length of the shortest path between these two points while remaining on the surface $S$.\\

This definition allows us to cite three cases $c_k$ according to the values of $k$:\\
\begin{itemize}\setlength{\itemsep}{4mm}
\item If  $k=0$ then  $c_k$  tends towards the reference point $p_0$ : $c_k = \{ p_0 \}$.
\item If  $k\rightarrow\propto$ then  $c_k$  is empty : $c_k=\{ 0\}$.
\item If  $0<k<\propto$ then  $c_k$  approaches  $S$ : $c_k =\{p\in S \backslash F(p,p_0) = k \}$.
\end{itemize}
\vspace{0.5cm}
\subsection{Riemannian analysis of facial curves}

To analyze the facial surfaces, we simply analyze the iso-geodesic curves that characterize these 3D face surfaces and compute a geodesic distance between them on a manifold depends on the Riemannian metric. To analyze the curve shape, we use the parameterization by the mathematical function SRVF (Square Root Velocity Function) [30, 31].\\

Let a parameterized closed curve be denoted as $\beta: I \rightarrow R^3$, for a unit interval $I \equiv [0, 2\pi]$, $\beta$ is represented by its $SRVF: q: I \rightarrow R^3$ defined as follow:

\begin{equation}\label{notre}
  q(s)= \frac{\beta'(s)}{(\parallel\beta(s)\parallel)^{\frac{1}{2}}}= \frac{\frac{d\beta(s)}{ds}}{\sqrt{\parallel\frac{d\beta(s)}{ds}\parallel}} \in R^3
\end{equation}
Where,
\begin{itemize}\setlength{\itemsep}{4mm}
\item $s \in I \equiv [0, 2\pi]$.
\item $\|.\|$ is the standard Euclidean norm in $R^3$.
\item $\|q(s)\|$  is the square-root of the instantaneous speed on the curve $\beta$.
\item $\frac{q(s)}{\|q(s)\|}$  is the instantaneous direction at the point $s \in [0, 2\pi]$ along the curve.
\end{itemize}
\vspace{0.5cm}
Thus, the curve $\beta$ can be recovered within a translation, using:
\begin{equation}\label{notre}
  \beta(s)= \int_{0}^{s} q(t)\|q(t)\| \, \mathrm{d}t
\end{equation}

We define the set of closed curves $\beta$ in $R^3$ by:
\begin{equation}\label{notre}
  C=\{q : S^1 \rightarrow R^3  \mid \int_{S^1}q(t)\parallel q(t)\parallel dt=0 \} \subset L(S^1,R^3)
\end{equation}
Where,
\begin{itemize}\setlength{\itemsep}{4mm}
\item $L(S^1,R^3)$. denotes the set of all functions integral $S^1$ to $R^3$
\item $\int_{S^1}q(t)\parallel q(t)\parallel dt$ denotes the total displacement in $R^3$ while moving from the origin of the curve until the end. When $\int_{S^1}q(t)\parallel q(t)\parallel dt=0$ the curve is closed.
\end{itemize}
\vspace{0.5cm}
All 3D closed curves are defined as nonlinear variety in the Helbert space. To analyze the shapes of the iso-geodesic curves and compute a geodesic distances between them, it is important to understand all vectors of their tangent spaces and impose a Riemannian structure. We equip the space of the closed curves of a Riemannian structure using the inner product defined as follows:
\begin{equation}\label{notre}
  <f,g>= \int_{0}^{1} (f(s),g(s)) ds
\end{equation}

Here, $f$ and $g$ are two vectors in the tangent space $T_v(c)$. We can also define $T_v(c)$:

\begin{equation}\label{notre}
 T_v(C)= \{ f:S^1 \rightarrow R^3 \mid <f(s),h(s)>=0, \quad  h \in N_v(c) \}
\end{equation}
$N_v(c)$ is a space of the normal vectors to the face curve.

After a mathematical representation of iso-geodesic curves using Riemannian metric, this metric should invariant certain transformation (translation, rotation, scale) [30]. The question to ask is how to compute the geodesic distance between two closed curves?. To answer this question, we used the approach introduced by Klassen in 2007 [31]. This method use path straightening flows to find a geodesic between two shapes.\\

To compare two facial surfaces, we just compare a pairs of closed curves of these two facial surfaces. Lets $c_1$ and $c_2$ two facial curves (iso-geodesic curves), $q_1$ and $q_2$ are respectively there Square Root Velocity Function ($SRVF$). The geodesic distance between $c_1$ and $c_2$ is computed by:

\begin{equation}\label{notre}
 d(q_1,q_2)= \int_0^1 \sqrt{<\varepsilon\prime(t),\varepsilon\prime(t)>}  dt
\end{equation}

With $\varepsilon$ is a geodesic path determined by the training method, this method is to connect the two curves by an arbitrary path $\alpha$ then update the path repeatedly in the negative direction of the gradient of the energy given by:

\begin{equation}\label{notre}
 E[\alpha] = \frac{1}{2} \int_0^1 <\frac{d}{ds}\alpha(t),\frac{d}{ds}\alpha(t)> dt
\end{equation}

$\varepsilon$ has been shown that the critical points of the energy equation $E[\alpha]$ are geodesic paths in $S$ [31, 13].\\

The facial surfaces $S_1$ and $S_2$ are represented by their iso-geodesic curves collection respectively $\{c_k^1; k \in [0,k_0]\}$ and $\{c_k^2; k \in [0,k_0]\}$, $k$ is a geodesic distance between $p_0$ (reference point) and $p$ two points of facial surface $S$.  The vectors of geodesic distances computed between a pairs of facial curves are used as input vectors of classification algorithms of our automatic facial recognition system.\\

To realize our 3D face recognition system, we use three classification algorithms are: the Neural Networks (NN), k-Nearest Neighbor (KNN) and Support Vector Machines (SVM).

\section{Simulation Results}

In this section we make a series of simulation to evaluate the effectiveness of the proposed approach. These results were performed based on SHREC 2008 database. This database contains total of 427 scans of 61 subjects (45 males and 16 females), for each of these 61 subjects 7 different scans, namely two “frontal”, one “look-up”, one “look-down”, one “smile”, one “laugh” and one “random expression” [21, 22].\\

In this paper, the features were extracted using Iso-Geodesic Curves (I-GC). This method was based on two principal steps: Iso-Geodesic Curves extraction using Fast Marching algorithm as solution of Eiconal equation and compute the length of the geodesic path between each facial curve and its corresponding curve using a Riemannian framework.\\

The following figure (6) summarizes the first experimental results of our simulation:

\begin{figure}[H]
\centering
\includegraphics[width=3.4in]{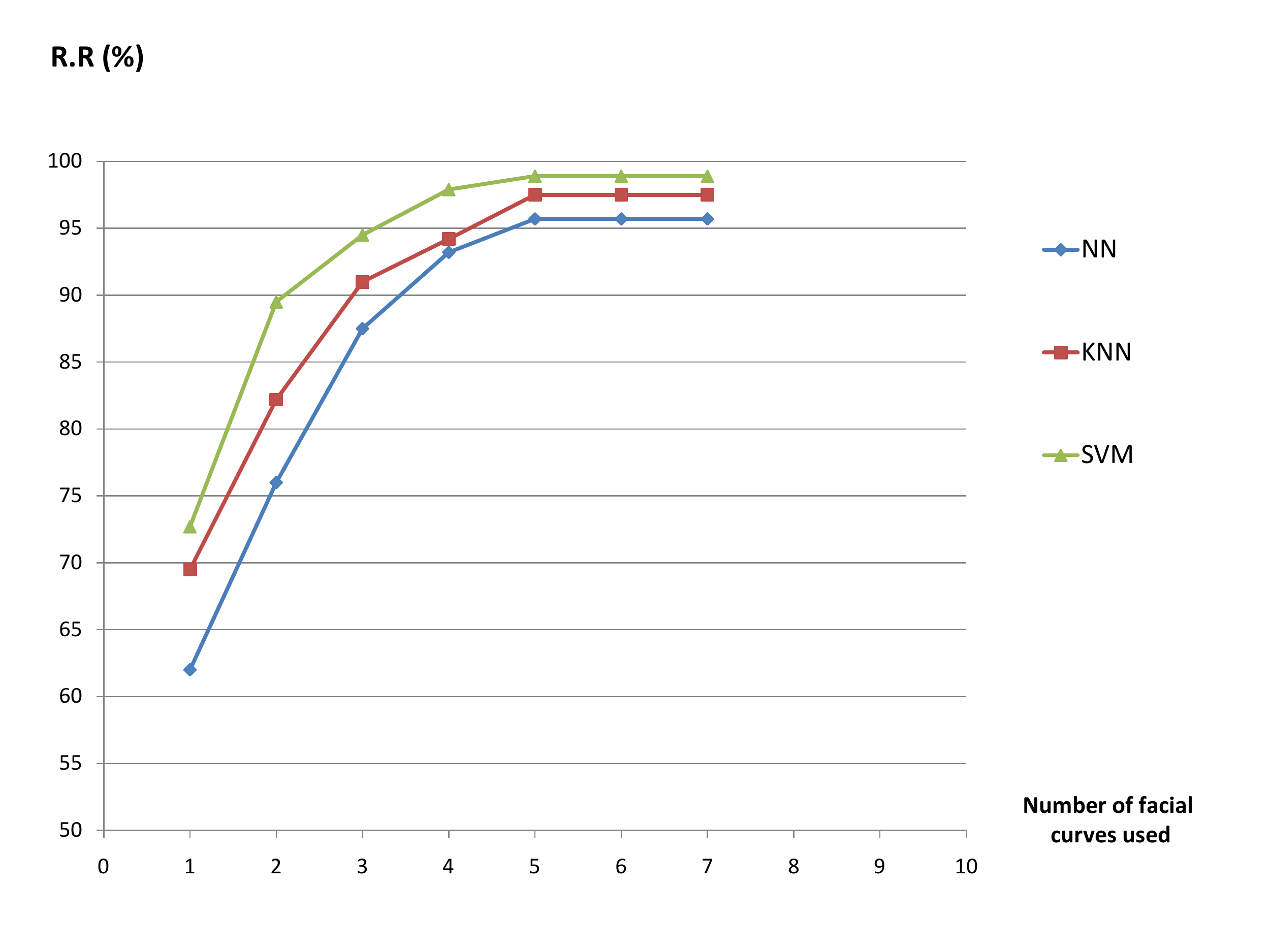}
\caption{Recognition rate in terms of number of facial curves used to represent the human face of SHREC 2008 database for each classification algorithm (NN, KNN and SVM)}
\label{fig_sim}
\end{figure}

Figure 5 shows the recognition rate in terms of number of facial curves used to represent a 3D human faces used in our systems. This figure shows that the images of the SHREC 2008 database are represented using five facial curves.\\

Given a candidate 3D face image ($Img$) of SHREC 2008 database. $Img$ is represented using five iso-geodesic curves. The shortest path between two 3D face images id defined as the sum of the distance between all pairs of corresponding facial curves in the two face images. The feature vector is then formed by the geodesic distances computed on all the curves and its dimension is equal to the number of used iso-geodesic curves (five curves for SHREC 2008 database). These vectors are used as input of classification algorithms of our 3D face recognition systems.

\begin{figure}[H]
\centering
\includegraphics[width=3in]{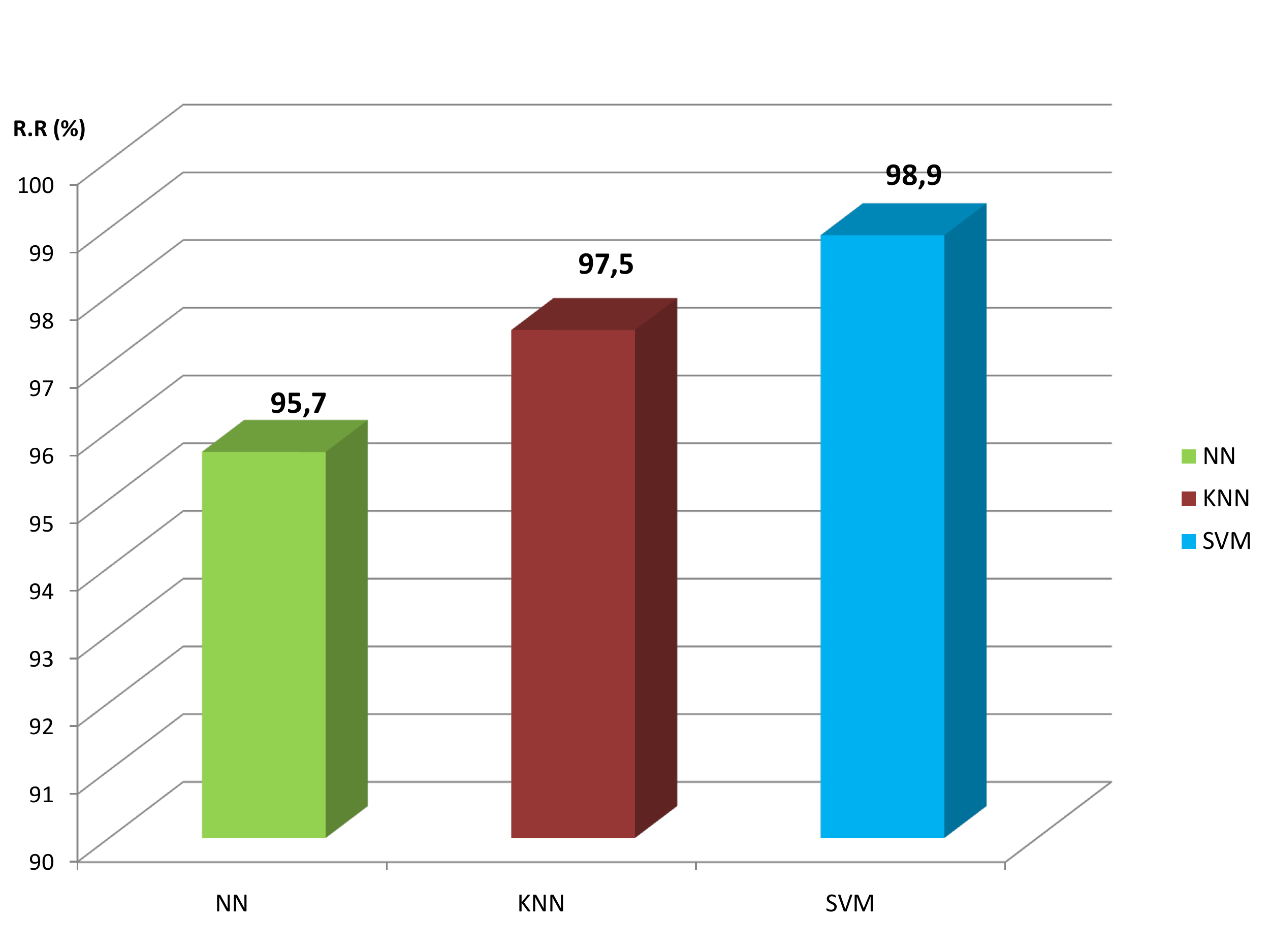}
\caption{Recognition Rate for SHREC 2008 images using three classification algorithms (NN, KNN and SVM)}
\label{fig_sim}
\end{figure}

Figure 7 shows the recognition rate for SHREC 2008 images using three classification algorithms: Neural Networks (NN), K-Nearest Neighbor (KNN) and Support Vector Machines (SVM). The best recognition rate was obtained using Support Vector Machines (SVM) as classification algorithm with recognition rate equal $98,9\%$.\\

In conclusion of this series of results, a summary table (Table2) compares the performance of our face authentication with respect to the performance obtained in other work systems.
\begin{table}[!h]
\centering
\begin{tabular}{|l|p{1.3cm}|p{2.7cm}|c|p{1.3cm}|}
\hline
\bf Date & \bf Reference & \bf Method & \bf Database & \bf Reported performance\\
\hline \hline
2004 & Haar et al [26] & facial Contour Curves & SHREC’08 & 91.1 \%  \\
\hline
2007 & Feng et al [24] & Euclidean Integral Invariants Signature (Closed 3D Curves) & FRGCv2 & 95,0\% \\
\hline
2007 & Samir et al [23] & Planar Curves Levels & Notre Dame & 90,4 \% \\
\hline
2007 & Samir et al [23] & Planar Curves Levels & FSU & 92,0\% \\
\hline
2008 & Daoudi et al [25] & Elastic Deformation Of Facial Surfaces (open paths) & FSU & 92,0 \% \\
\hline
2015 & Ahdid et al [29] & GD+LDA+SVM & SHREC’08 & 93,2\% \\
\hline
2015 & Ahdid et al [29] & GD+PCA+SVM & SHREC’08 & 95,3 \% \\
\hline
\bf 2016 & \bf Our approach & \bf Iso-Geodesic Curves + SVM & \bf SHREC’08 & \bf 98,9\% \\
\hline
\end{tabular}
\caption{COMPARISON OF OUR METHOD WITH OTHER METHODS OBTAINED IN OTHER WORK SYSTEMS}
\end{table}

We can notice that the performance of our automatic 3D face recognition system, In addition our system is perfect in all assessment. Our goal was to improve 3D faces recognition system we affirm based on the results that our goal is achieved.

\section{Conclusion}
In this work, we have presented a 3D face recognition system based on the computation of the geodesic distance between the reference point and the other points in the 3D face surface. This method represents a face surface as a collection of Iso-Geodesic curves and computes a geodesic distance between each pairs of these facial curves. For the classifying step we have implemented algorithms as Neural Networks (NN), K-Nearest Neighbor (KNN) and Support Vector Machines (SVM). Simulation results show us a better recognition rate ($98.9\%$) using SVM classification algorithm.

\end{document}